\documentclass[10pt,twocolumn,letterpaper]{article}

\usepackage{cvpr}
\usepackage{times}
\usepackage{epsfig}
\usepackage{graphicx}
\usepackage{amssymb}
\usepackage{multicol}
\usepackage{multirow}
\usepackage{stfloats}
\usepackage{amsthm}
\usepackage{empheq}
\usepackage{amsmath}
\usepackage{multicol}
\usepackage{multirow}
\usepackage[center]{caption}
\usepackage{pdfpages}

\usepackage[pagebackref=true,breaklinks=true,letterpaper=true,colorlinks,bookmarks=false]{hyperref}

\usepackage{etoolbox,siunitx}
\robustify\bfseries

\graphicspath{{./images/}}

\cvprfinalcopy %

\def\cvprPaperID{9739} %

\ifcvprfinal\fi
\begin{document}

\title{Training Deep SLAM on Single Frames}
\author{Igor Slinko, Anna Vorontsova, Dmitry Zhukov, Olga Barinova, Anton Konushin \\
Samsung AI Center Russia, Moscow\\
{\tt\small \{i.slynko,a.vorontsova,d.zhukov,o.barinova,a.konushin\}@samsung.com}\\
}

\maketitle

\section*{Abstract}
\textit{
Learning-based visual odometry and SLAM methods demonstrate a steady improvement over past years. However, collecting ground truth poses to train these methods is difficult and expensive. This could be resolved by training in an unsupervised mode, but there is still a large gap between performance of unsupervised and supervised methods. In this work, we focus on generating synthetic data for deep learning-based visual odometry and SLAM methods that take optical flow as an input. We produce training data in a form of optical flow that corresponds to arbitrary camera movement between a real frame and a virtual frame. For synthesizing data we use depth maps either produced by a depth sensor or estimated from stereo pair. We train visual odometry model on synthetic data and do not use ground truth poses hence this model can be considered unsupervised. Also it can be classified as monocular as we do not use depth maps on inference.}

\textit{We also propose a simple way to convert any visual odometry model into a SLAM method based on frame matching and graph optimization. We demonstrate that both the synthetically-trained visual odometry model and the proposed SLAM method build upon this model yields state-of-the-art results among unsupervised methods on KITTI dataset and shows promising results on a challenging EuRoC dataset.}

\section{Introduction}

Simultaneous localization and mapping is an essential part of robotics and augmented reality systems. Deep learning-based methods for visual odometry and SLAM~\cite{wang2017deepvo, Wang2018Espvo, zhou2018deeptam, Henriques2018MapNet, Xue019BeyondTracking} have evolved over the last years and are able to compete with classical geometry-based methods on the datasets with predominant motions along plain surfaces like KITTI\cite{Geiger2012CVPR} and Malaga\cite{blanco2009cor}. 

\begin{figure}[ht]
  \includegraphics[width=\linewidth]{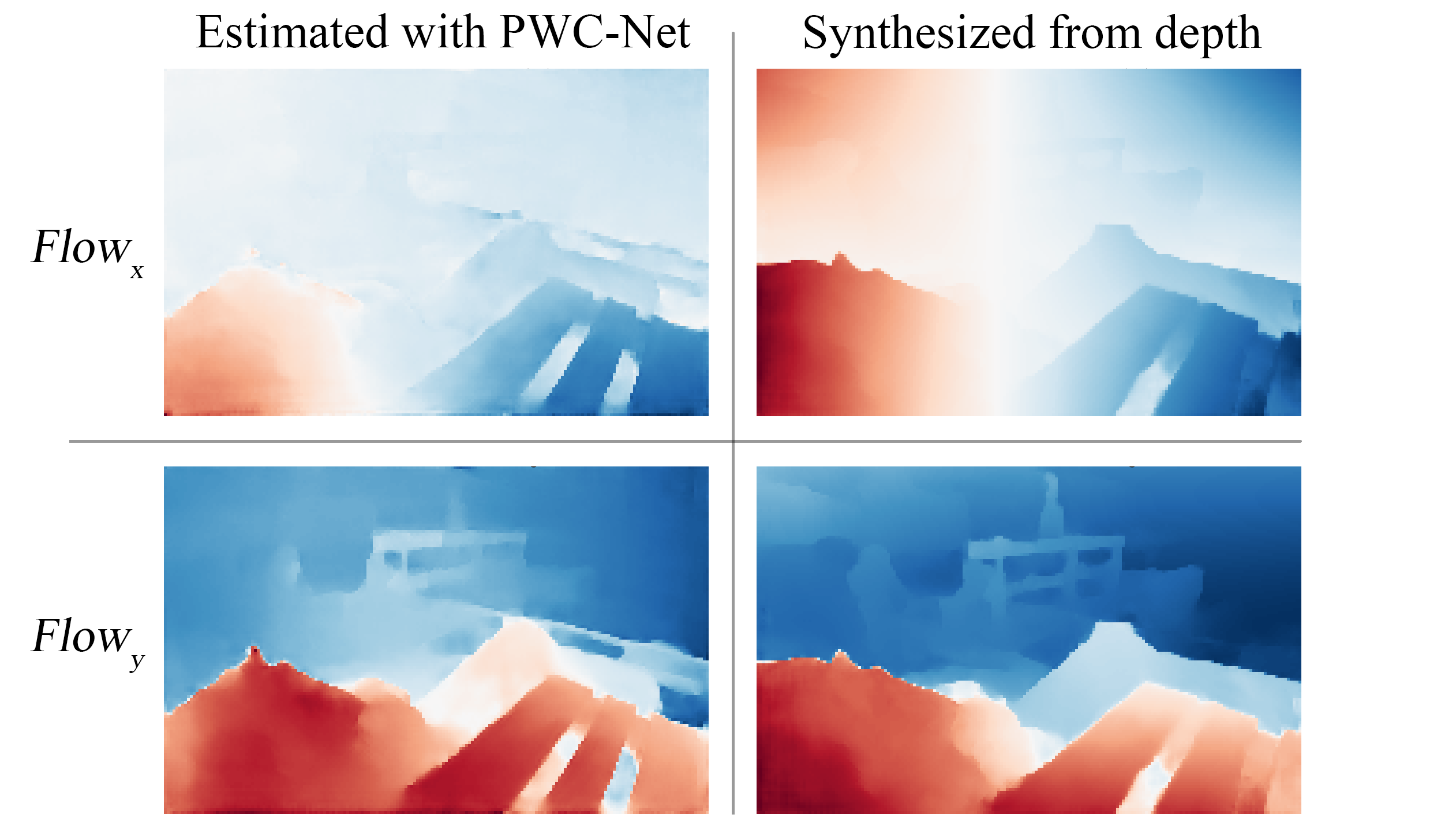}
  \caption{Optical flow estimated with trainable method (PWC-Net) on consecutive frames against the one synthesized from depth map and given 6DoF.}
  \label{fig:euroc_synthetic_flow}
\end{figure}

However, the practical use of deep learning-based methods for visual odometry and SLAM is limited by the difficulty of acquiring precise ground truth camera poses. To overcome this problem, unsupervised visual odometry methods are being actively investigated~\cite{almalioglu2018ganvo, Feng2019sganvo, Zhan2018Depth-VO-Feat, Li2017UnDeepVO, yin2018geonet}. Existing methods~\cite{zhou2018deeptam, zhou2017unsupervised, Teed2019DeepV2D, DBLP:journals/corr/Vijayanarasimhan17, bian2019sc-sfmlearner} use video sequences for training. They estimate camera movements and depth maps jointly. For a pair of consecutive frames, the first frame is re-rendered from the point of view of the second one. The difference between re-rendered first frame and ground truth second frame is used in a loss function.

We propose an alternative approach, that requires only individual frames with depth rather than video sequences. In our approach, we sample random camera motions with respect to the physical motion model of an agent. Then for each sampled camera motion we synthesize corresponding optical flow between a real frame and a virtual frame. The resulting synthesized optical flow with generated ground truth camera motions can be used for training a learning-based visual odometry model.

Our contribution is twofold: \begin{itemize}
    \item First, we introduce an unsupervised method of training visual odometry and SLAM models on synthetic optical flow generated from depth map and arbitrary camera movement between selected real frame and virtual frame. This approach does not use frame sequences for training and does not require ground truth camera poses.
    \item Second, we propose a simple way to convert any visual odometry model into a SLAM system with frame matching and graph optimization.
\end{itemize}

We demonstrate that our approach outperforms state-of-the-art unsupervised deep SLAM methods on KITTI dataset. Also we tried our method on challenging EuRoC dataset, and to the best of our knowledge, this is the first unsupervised learnable method ever evaluated on EuRoC.  

\section{Related work}
\subsection{Classical methods}

Several different mathematical formulations for visual odometry have been considered in the literature. Geometry-based visual odometry methods can be classified into direct (\eg~\cite{kerl2013robust}) and indirect (\eg~\cite{mur2017orb}) or dense (\eg~\cite{steinbrucker2011real}) and sparse (\eg~\cite{engel2018direct}).

Direct methods take original images as inputs, while indirect methods process detected keypoints and corresponding features. 
Dense methods accept regular inputs such as images, optical flow or dense feature representations. In sparse methods, data of irregular structure is used. 

Many of the classical works apply bundle adjustment or pose graph optimization in order to mitigate the odometry drift. Since this strategy showed its effectiveness in related tasks~\cite{mur2017orb, Schops2019badslam}, we adopt it in our deep learning-based approach.

\subsection{Supervised learning-based methods}

DeepVO~\cite{wang2017deepvo} was a pioneer work to use deep learning for visual odometry. This deep recurrent network regresses camera motion using pretrained FlowNet~\cite{dosovitskiy2015flownet} as a feature extractor. ESP-VO~\cite{wang2018end} extends this model with sequence-to-sequence learning and introduces an additional loss on global poses. LS-VO~\cite{costante2018ls} also uses the result of FlowNet and formulates the problem of estimating camera motion as finding a low-dimensional subspace of the optical flow space. DeMoN~\cite{ummenhofer2017demon} estimates both camera motion, optical flow and depth in the EM-like iterative network. By efficient usage of consecutive frames, DeMoN improves accuracy of depth prediction over single-view methods. This work became a basis for the first deep SLAM method called DeepTAM~\cite{zhou2018deeptam}. Similar ideas were implemented in ENG~\cite{dharmasiri2018eng}, which was proved to work on both indoor and outdoor datasets.

\subsection{Unsupervised learning-based methods}

Recent advances in simultaneous depth and motion estimation~\cite{zhou2018deeptam}~\cite{Teed2019DeepV2D} from video sequences allow to track more accurate camera position in an unsupervised manner. These methods exploit sequential nature of the data in order to model scene dynamics and take clues from occlusion, between-frame optical flow consistency and other factors (\cite{zhou2017unsupervised},~\cite{DBLP:journals/corr/abs-1901-07288},~\cite{DBLP:journals/corr/Vijayanarasimhan17}). To achieve motion consistency, additional modalities of data such as depth maps are estimated in a joint pipeline. Similarly to these approaches, we use depth maps; however, we once estimate depth maps from a stereo pair, and keep them unchanged during optical flow synthesis. 

\subsection{Novel view synthesis}

The idea of novel view synthesis using single frame or stereo pair and optical flow has been exploited in~\cite{zhou2018deeptam} . In this method, model is trained in a supervised manner by minimizing difference between estimated and ground truth camera poses. Novel view synthesis is used as a part of working pipeline, with new camera position predicted, virtual frame synthesized, and movement between virtual and current frame estimated. Therefore, this method operates mainly in the image domain, utilizing optical flow only to transit between different image instances. In our approach, we synthesize optical flow rather than images. We also generate training data only on training stage, and do not use it during inference.  

\section{Proposed method}

\subsection{Visual odometry model}

\begin{figure*}[ht]
  \includegraphics[width=\linewidth]{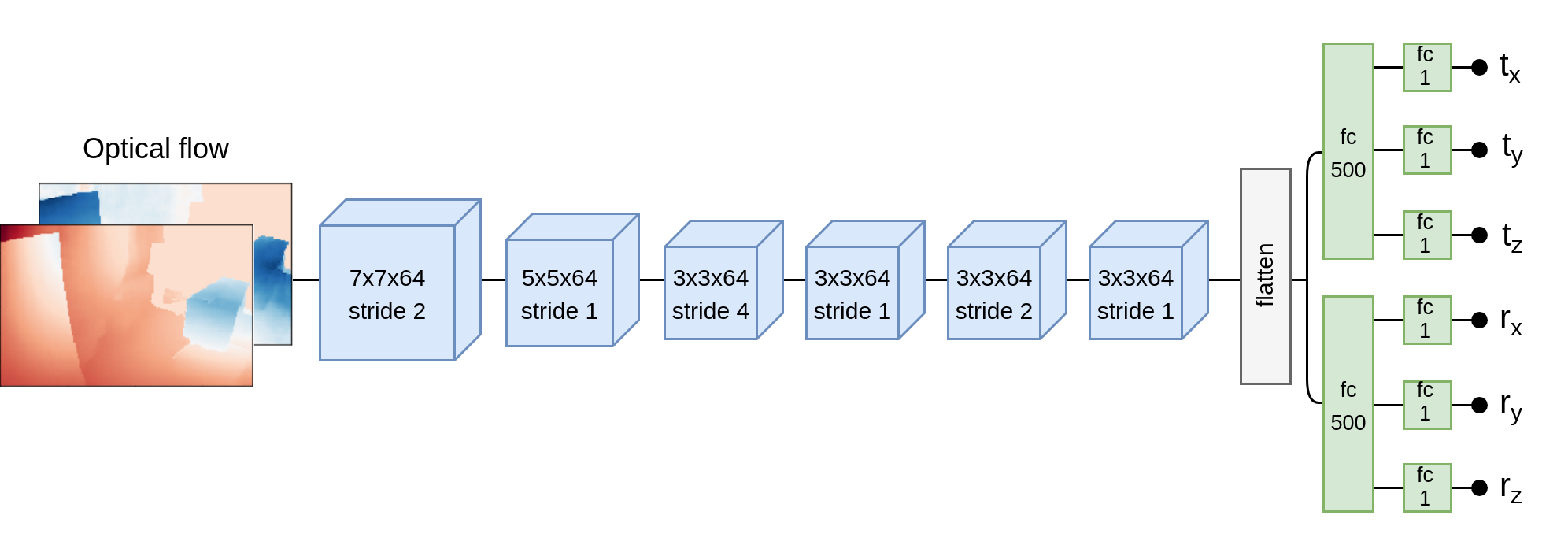}
  \caption{Architecture of visual odometry model}
  \label{fig:visual_odometry_network}
\end{figure*}

For visual odometry, we adopt a neural network from~\cite{Slinko2019MotionMap} that estimates relative rotation and translation in form of 6DoF from dense optical flow. The model architecture is shown on Figure \ref{fig:visual_odometry_network}. Generally speaking. our approach can be applied to any model that takes optical flow as an input and predicts 6DoF.

We use PWC-Net~\cite{sun2018pwc} for optical flow estimation. The source code and pretrained weights are taken from the official repository \footnote{\url{https://github.com/NVlabs/PWC-Net}}. For KITTI, we opted for weights pretrained on FlyingThings3D~\cite{MIFDB16} dataset and fine-tuned on KITTI optical flow dataset\cite{Menze2018JPRS}. For EuRoC, which is in grayscale, we fine-tuned PWC-Net weights and using Sintel dataset~\cite{Butler:ECCV:2012} converted to grayscale.

In our experiments, we found out that a single neural network can effectively handle motions within a certain range. However, motions between the first and the last frames in loops differ significantly from motions between consecutive frames: not only are they of different scale but also much more diverse. To address this problem, we train two models: $NN_{cons}$ model to estimate motions between consecutive frames and $NN_{loops}$ to predict motions specifically between the first and the last frames in a loop. 

\subsection{Synthetic training data generation}

Taking depth map and an arbitrary motion in form of 6DoF, the method runs as follows: 
\begin{enumerate}
    \item First, we map depth map pixels into points in a frustum that gives a point cloud. 
    \item To build a virtual point cloud, we represent motion in form of SE3 matrix and apply it to the current point cloud, obtaining this point cloud from another view point. 
    \item Next, we re-project this virtual point cloud back to image plane, that gives shifted pixel grid. 
    \item To get absolute values of optical flow, we calculate the difference between re-projection and regular pixel grid of a source depth map.
\end{enumerate}
Then, this newly synthesized optical flow can be used as an input to a visual odometry network. 

\begin{figure*}[ht]
  \includegraphics[width=\linewidth]{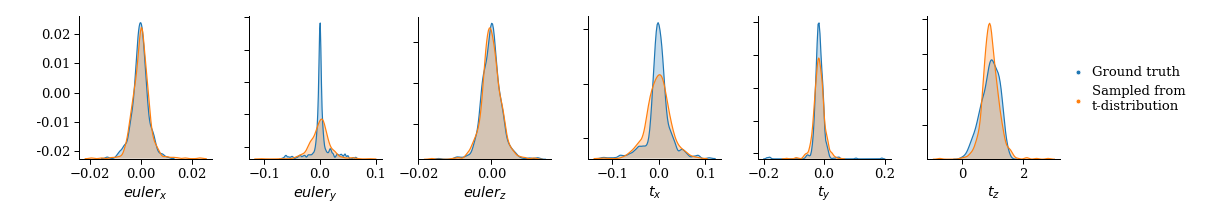}
  \caption{Distribution of 6DoF motion in KITTI dataset}
  \label{fig:dof_distribution}
\end{figure*}

The camera movements should be generated with taking physical motion model of an agent into account. Since existing datasets do not contain such information, we estimate the motion model from the ground truth data. By modelling, we approximate ground truth distribution of 6DoF using Student's t-distribution Figure \ref{fig:dof_distribution}. We adjust the parameters of this distribution once and keep them fixed during training, while 6DoF are being sampled randomly. 

In case that dataset does not contain depth maps, they can be estimated from a monocular image or a stereo pair. In our experiments, we obtain depth $z$ from disparity $d$ similar to~\cite{ZbontarL2015disparity}.
\begin{gather}
    z = \frac{f B}{d}, 
\end{gather}
where $f$ is focal length and $B$ is distance between stereo cameras.

To estimate disparity, we match left and right image with the same PWC-Net~\cite{sun2018pwc} as was used to estimate optical flow.

Since we do not need ground truth camera poses for data synthesis, the proposed approach can be considered unsupervised.

\begin{figure*}[ht]
  \includegraphics[width=\linewidth]{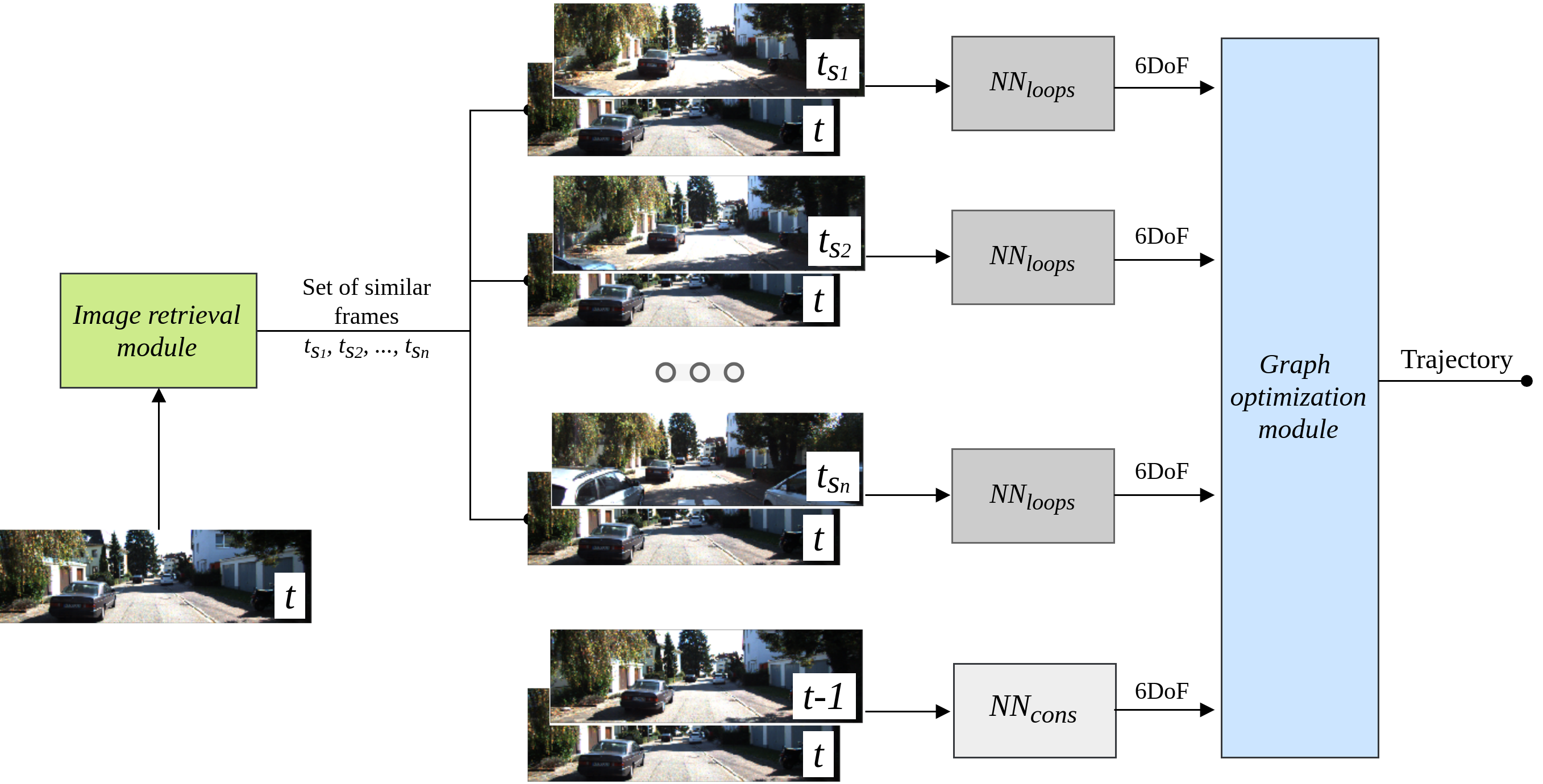}
  \caption{Architecture of proposed SLAM method}
  \label{fig:slam}
\end{figure*}

\subsection{Relocalization}

Relocalization can be reformulated as image retrieval task Figure \ref{fig:slam}. Following a standard approach, we measure distance between frames according to their visual similarity. Here, we use classical Bag of Visual Words (BoVW) from OpenCV library~\cite{opencv_library} that is applied to SIFT features~\cite{Lowe1999Sift}. These features are stored in a database. To create a topological map, for each new frame its 20 nearest neighbors are extracted from database. These found frames are further filtered by applying Lowe's ratio test~\cite{Lowe2004Test} and rejecting candidates with less then $N_{th}$ matched keypoints.  

\subsection{Graph optimization}

We adopt graph optimization in order to expand a visual odometry method to a SLAM algorithm. One way to formulate SLAM is to use a graph with nodes corresponding to the camera poses and edges representing constraints between these poses. Two camera poses are connected with an edge if they correspond to consecutive frames or if they are considered similar by relocalization module. The edge constraints are obtained as relative motions predicted with visual odometry module. Once such a graph is constructed, it can be further optimized in order to find the spatial configuration of the nodes that is the most consistent with the relative motions modeled by the edges. The nodes obtained through optimization procedure are then used as final pose estimates. The reported metrics are thus computed by comparing these estimates with ground truth poses.

We opted for a publicly available g2o library~\cite{kuemmerle2011g2o} that implements least-squares error minimization. To incorporate optimization module in our Python-based pipeline we use Python binding for g2o \footnote{\url{https://github.com/uoip/g2opy}}.

The interaction between visual odometry networks $NN_{cons}$, $NN_{loops}$ and graph optimization module is guided by a set of hyperparameters: \begin{itemize}
    \item $C_{s_i}$ -- coefficient for standard deviation
    \item $C_r$ -- extra scaling factor for rotation component
    \item $T_{loop}$ -- loop threshold: a loop is detected if difference between indices of two images exceeds given threshold. In this case, relative motion is predicted using loop network $NN_{loops}$, otherwise $NN_{cons}$ is used.
\end{itemize}

We adjust these hyperparameters on a validation subset and then evaluate our method on a test subset.

To construct a graph, we need to pass $7 \times 7$ information matrix $P_i$ corresponding to 3D translation vector and rotation in form of a quaternion. 

First, we compose resulting covariance matrix as: 
\begin{gather}
    Q_i = C_{s_i}
    \begin{bmatrix}
        \sigma_{t_x}^2 & 0 & 0 & 0 & 0 & 0 \\
        0 & \sigma_{t_y}^2 & 0 & 0 & 0 & 0 \\
        0 & 0 & \sigma_{t_z}^2 & 0 & 0 & 0 \\
        0 & 0 & 0 & C_r \sigma_{\alpha}^2 & 0 & 0 \\
        0 & 0 & 0 & 0 & C_r \sigma_{\beta}^2 & 0 \\
        0 & 0 & 0 & 0 & 0 & C_r \sigma_{\gamma}^2 \\
    \end{bmatrix}
\end{gather}
where $\alpha, \beta, \gamma$ stand for Euler angles $euler_x, \ euler_y, \ euler_z$ respectively.

A conversion between this matrix and $Q_i$ is performed according to~\cite{Claraco2010Tutorial}:
\begin{gather}
    P_i = \left( \frac{\partial p_7 (p_6)}{\partial p_6} Q_i \frac{\partial p_7 (p_6)}{\partial p_6} \right)^{-1}
\end{gather}

\section{Experiments}

\subsection{Datasets}

\noindent \textbf{KITTI odometry 2012.} We used KITTI dataset to evaluate our method in a simple scenario. We trained on trajectories 00, 02, 08 and 09 and tested on trajectories 03, 04, 05, 06, 07, 10.

It is worth noting that ground truth poses were collected using GPS sensor that yielded noisy measurements of motion along y-axis (there are vertical movements perpendicular to the surface of the road). This effect is cumulative: while relative motion between consecutive frames was measured quite precisely, the difference in absolute height between the first and the last frame of a loop may be up to 2 meters. 

\noindent \textbf{EuRoC.} This dataset was recorded using flying drone in two different environments. 6DoF ground truth poses are captured by laser tracker or motion capture system depending on environment. The sensor and ground truth data are calibrated and temporally aligned, with all extrinsic and intrinsic calibration parameters provided. As original frames come unrectified, preprocessing included removing distortion. We validated on trajectories MH\_02\_easy, V1\_02\_medium and tested on V1\_03\_difficult and V2\_03\_difficult, while other trajectories were used for training.

Due to complexity of the environment, dynamic motions and weakly correlated, entangled trajectories, EuRoC appears to be a challenging task for trainable methods. Moreover, images are in grayscale, that adds difficulty for methods that match pixels based on their color rather than pure intensity. To the best of our knowledge, we present the first trainable method that demonstrates competitive results on EuRoC among all methods trained in an unsupervised manner. 

\begin{figure*}[!ht]
  \includegraphics[width=\linewidth]{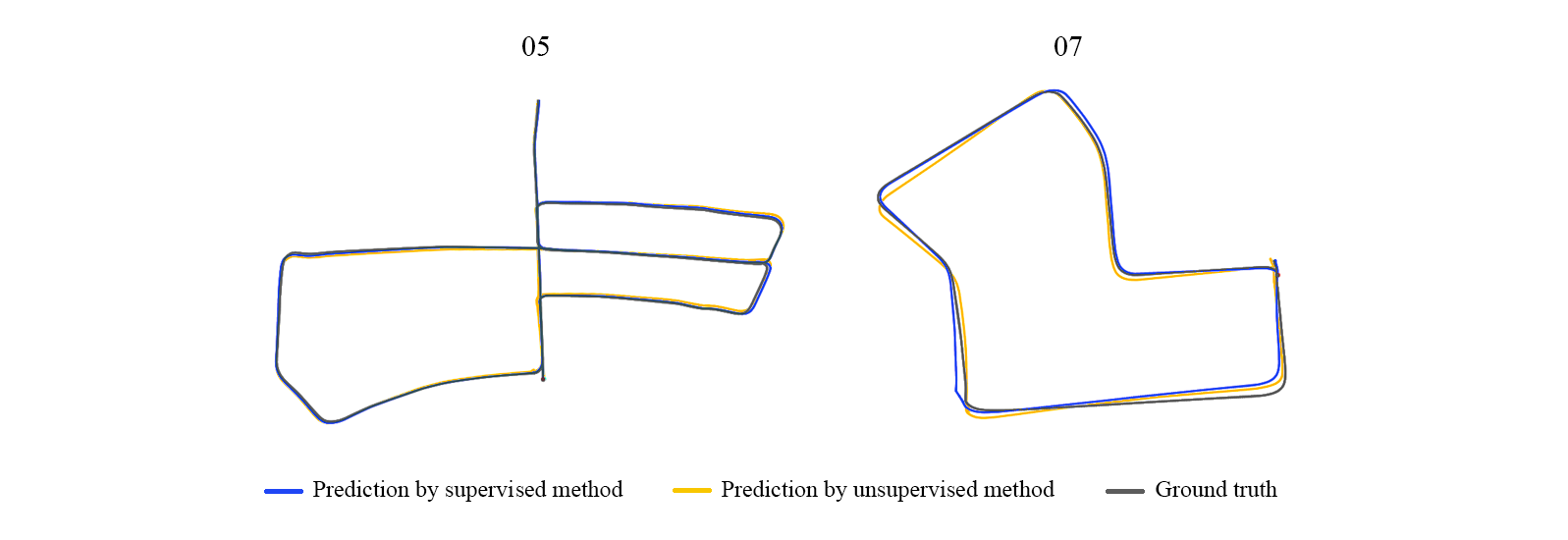}
  \caption{Ground truth and estimated KITTI trajectories}
  \label{fig:kitti_trajectories}
\end{figure*}

\begin{table}
     \begin{tabular}{|c|c|c|c|}
     \hline
     Method & ATE & $t_{err}$ & $r_{err}$  \\
     \hline
     $NN_{cons}$ & $4.38 \pm 0.36$ & $2.07 \pm 0.03$ & $0.93 \pm 0.04$ \\
     \hline
     $NN_{loops}$ & $9.33 \pm 1.04$ & $3.15 \pm 0.12$ & $1.42 \pm 0.03$ \\
     \hline
     SLAM & $1.84 \pm 0.06$ & $1.54 \pm 0.04$ & $0.74 \pm 0.04$ \\
     \hline
    \end{tabular}
    \caption{
        Results of supervised visual odometry models and SLAM method on KITTI dataset. Optimal parameters for SLAM are $C_{s_i}=10000$, $C_r=1$, $T_{loop}=50$ 
    }
    \label{table:kitti_supervised}
\end{table}

\begin{table}
     \begin{tabular}{|c|c|c|c|} 
     \hline
     Method & ATE & $t_{err}$ & $r_{err}$  \\ 
     \hline
     $NN_{cons}$ & $14.30 \pm 1.57$ & $5.57 \pm 0.33$ & $2.23\pm 0.15$ \\ 
     \hline
     $NN_{loops}$ & $16.21 \pm 1.42$ & $6.43 \pm 0.23$ & $2.82 \pm 0.10$ \\
     \hline
     SLAM & $3.24 \pm 0.17$ & $3.37 \pm 0.12$ & $1.24 \pm 0.04$ \\
     \hline
    \end{tabular}
    \caption{
        Results of unsupervised visual odometry models and SLAM method on KITTI dataset. Optimal parameters for SLAM are $C_{s_i}=10000$, $C_r=0.004$, $T_{loop}=50$}
    \label{table:kitti_unsupervised_trainable}
\end{table}

\begin{small}
    \begin{table}
    \begin{center}
     \begin{tabular}{|c|c|c|c|} 
     \hline
     Method & $t_{err}$ & $r_{err}$ \\ 
     \hline
     ORB-SLAM2\cite{mur2017orb} & 2.41 & \textbf{0.245} \\
     \hline
     Ours, SLAM & \textbf{1.54} & 0.74 \\
     \hline
    \end{tabular}
    \caption{
        Metrics on KITTI for supervised methods. \\ Numbers are taken from~\cite{Zhan2019learnt}
    }
     \label{table:kitti_supervised_all}
    \end{center}
    \end{table}
\end{small}

\begin{small}
    \begin{table}
     \centering
     \begin{tabular}{|l|c|c|c|} 
     \hline
     Method & ATE & $t_{err}$ & $r_{err}$ \\ 
     \hline
     SfMLearner\cite{zhou2017unsupervised} & 28.14 & 12.21 & 4.74 \\
     \hline
     Depth-VO-Feat\cite{Zhan2018Depth-VO-Feat} & 16.83 & 8.15 & 4.00 \\
     \hline
     SC-SfMLearner\cite{bian2019sc-sfmlearner} & 17.92 & 7.42 & 3.35 \\
     \hline
     UnDeepVO\cite{Li2017UnDeepVO} & – & 6.27 & 3.39 \\
     \hline
     GeoNet\cite{yin2018geonet} & – & 13.12 & 7.38 \\
     \hline
     Vid2Depth\cite{mahjourian2018unsupervised} & – & 37.98 & 18.24 \\
     \hline
     SGANVO\cite{Feng2019sganvo} & – & 5.12 & 2.53 \\
     \hline
     Ours, VO & 14.30 & 5.57 & 2.23 \\ 
     \hline
     Ours, SLAM & \textbf{~~3.24} & \textbf{~~3.37} & \textbf{~~1.24} \\
     \hline
    \end{tabular}
    \caption{
        Metrics on KITTI for unsupervised methods. \\ Numbers are taken from~\cite{Zhan2019learnt}
    }
    \label{table:kitti_unsupervised_all}
    \end{table}
\end{small}

\subsection{Training procedure}

The visual odometry model is trained from scratch using Adam optimization algorithm with amsgrad option switched on. The batch size is set to 128, the momentum is fixed to (0.9, 0.999). 

In our experiments, we noticed that despite loss being almost constant among several re-runs, final metrics (ATE, RPE etc.) may fluctuate significantly. This spreading is assumed to be caused by optimization algorithm terminating at different local minima depending on weights initialization and randomness incorporated by sampling batches. We address this challenge by adopting learning rate schedule in order to force optimization algorithm to traverse several local minima during the training process. Switching our training procedure for cyclic learning rate helped to decrease standard deviation of final metrics and the values of metrics themselves.

Initially, values of learning rate are bounded by [~0.0001,~0.001~]. In addition, if validation loss does not improve for 10 epochs, both the lower and upper bounds are multiplied by 0.5. Training process is terminated when learning rate becomes negligibly small. We used $10^{-5}$ as a learning rate threshold. Under these conditions, models are typically trained for about 80 epochs.

In several papers on trainable visual odometry~\cite{costante2018ls, Lv18eccv, wang2017deepvo, zhao2018learning, zhou2018deeptam}, different weights are used for translation loss and rotation loss. Since small rotation errors may have a crucial impact on the shape of trajectory, precise estimation of Euler angles is more important compared to translations. We multiply loss for rotation components by 50, as it was proposed in~\cite{costante2018ls}.

\subsection{Evaluation protocol} 

We evaluate visual odometry methods with several commonly used metrics.

For KITTI, we follow the evaluation protocol implemented in KITTI devkit \footnote{\url{https://github.com/alexkreimer/odometry/devkit}} that computes translation ($t_{err}$) and rotation ($r_{err}$) errors. Both translation and rotation errors are calculated as root-mean-squared-error for all possible sub-sequences of length (100,~\dots,~800) meters. The metrics reported are the average values of these errors per 100 meters.

For EuRoC, we use RPE metric that measures frame-to-frame relative pose error. 

To provide a detailed analysis, we also report values of absolute trajectory error (ATE), that measures the average pairwise distance between predicted and ground truth camera poses.

Since the results between different runs vary significantly, in order to obtain fair results we conduct all experiments for 5 times with different random seeds. The metrics reported are mean and standard deviation of execution results.

\subsection{Results on KITTI}

The results of our supervised and unsupervised visual odometry and SLAM are listed in Tab.~\ref{table:kitti_supervised} and Tab.~\ref{table:kitti_unsupervised_trainable}, respectively. According to them, visual odometry network $NN_{cons}$ trained on consecutive frames yields better results comparing to $NN_{loops}$ trained to estimate targets coming from a wider distribution. Combination of these two networks within a deep SLAM architecture helps to improve accuracy of predictions significantly. 

The existing quality gap between supervised and unsupervised approaches can be explained by non-rigidity of the scene, that exceeds the limitations of our data generation method. To obtain synthetic optical flow, a combination of translation and rotation is applied to a point cloud. Since it does not affect pairwise distances between points, the shapes of objects presenting in the scene remain unchanged and no new points appear. Thereby, rigidity of the scene is implicitly incorporated into the data synthesizing pipeline. For KITTI, scene does not meet these requirements due to the large displacements between consecutive frames and numerous moving objects appearing in the scene. 

According to Tab.~\ref{table:kitti_supervised_all}, the proposed method is comparable with ORB-SLAM2. We summarize results of unsupervised learnable methods in Tab.~\ref{table:kitti_unsupervised_all}. We show that our method significantly outperforms current state-of-the-art among all unsupervised deep learning-based approaches to trajectory estimation. 
\begin{figure}[!t]
  \includegraphics[width=\linewidth]{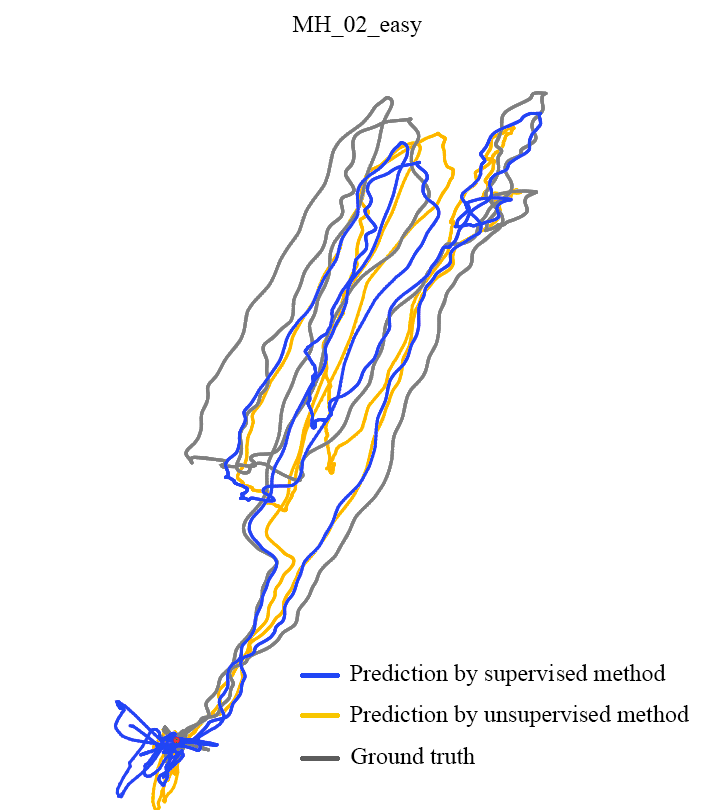}
  \caption{Ground truth and estimated EuRoC trajectory}
  \label{fig:euroc_trajectories}
\end{figure}

\subsection{Results on EuRoC}
For EuRoC dataset, we observed that it is more profitable to train visual odometry $NN_{cons}$ on a mixture of strides 1, 2, 3, rather than training on a single stride. Results of $NN_{cons}$ and $NN_{loops}$ are presented in Tab.~\ref{table:euroc_supervised} for supervised training and in Tab.~\ref{table:euroc_unsipervised} for unsupervised training. We expected the results on EuRoC to resemble KITTI results, where supervised method surpasses unsupervised method remarkably. Surprisingly, metrics for our SLAM model trained in supervised and unsupervised manner are nearly the same. We attribute this to the following reasons. Firstly, since EuRoC scenes are rigid, generated flow looks similar to estimated flow. Secondly, randomly sampled training data prevent unsupervised method from overfitting, while supervised method tends to memorize the entire dataset.

\begin{table*}
    \begin{center}
     \begin{tabular}{|c|c|c|c|c|c|c|} 
     \hline
     Method & val ATE & val RPE\textsubscript{t} & val RPE\textsubscript{r} & test ATE & test RPE\textsubscript{t} & test RPE\textsubscript{r}  \\ 
     \hline
     $NN_{1+2+3}$ & $1.35 \pm 0.07$ & $3.16 \pm 0.20$ & $19.75 \pm 1.20$ & $1.32 \pm 0.06$ & $2.78 \pm 0.25$ & $55.76 \pm 2.16$ \\ 
     \hline
     $NN_{loops}$ & $1.43 \pm 0.11$ & $3.64 \pm 0.30$ & $23.88 \pm 4.78$ & $1.36 \pm 0.05$ & $3.02 \pm 0.13$ & $53.45 \pm 2.00$ \\
     \hline
     SLAM & $~~0.51 \pm 0.015$ & $1.06 \pm 0.03$ & $~~8.36 \pm 0.17$ & $0.81 \pm 0.01$ & $1.51 \pm 0.02$ & $19.59 \pm 1.37$\\
     \hline
     \end{tabular}
     \caption{
        Results of supervised visual odometry models and SLAM method on EuRoC dataset. \\ Optimal parameters for SLAM are $C_{s_i}=10000$, $C_r=0.001$, $T_{loop}=100$
    }
    \label{table:euroc_supervised}
    \end{center}
\end{table*}

\begin{table*}
    \begin{center}
     \begin{tabular}{|c|c|c|c|c|c|c|} 
     \hline
     Method & val ATE & val RPE\textsubscript{t} & val RPE\textsubscript{r} & test ATE & test RPE\textsubscript{t} & test RPE\textsubscript{r} \\ 
     \hline
     $NN_{1+2+3}$ & $1.06 \pm 0.04$ & $2.26 \pm 0.14$ & $20.97 \pm 1.39$ & $1.35 \pm 0.46$ & $3.04 \pm 0.35$ & $62.56 \pm 3.29$ \\
     \hline
     $NN_{loops}$ & $1.37 \pm 0.12$ & $4.23 \pm 0.59$ & $33.49 \pm 1.73$ & $1.28 \pm 0.05$ & $3.43 \pm 0.16$ & $67.17 \pm 4.36$ \\
     \hline
     SLAM & $~~0.57 \pm 0.008$ & $1.12 \pm 0.03$ & $~~9.03 \pm 0.20$ & $0.84 \pm 0.17$ & $1.49 \pm 0.27$ & $23.13 \pm 7.40$\\
     \hline
     \end{tabular}
     \caption{
        Results of unsupervised visual odometry models and SLAM method on EuRoC dataset. \\ Optimal parameters for SLAM are $C_{s_i}=1000$, $C_r=0.0001$, $T_{loop}=100$
    }
    \label{table:euroc_unsipervised}
    \end{center}
\end{table*}

\vspace*{2cm} $ $\\

\section{Conclusion}

We proposed an unsupervised method of training visual odometry and SLAM models on synthetic optical flow generated from depth map and arbitrary camera movement between selected real frame and virtual frame. This approach does not use frame sequences for training and does not require ground truth camera poses.

We also presented a simple way to build SLAM system from an arbitrary visual odometry model. To prove our ideas, we conducted experiments of training unsupervised SLAM on KITTI and EuRoC datasets. The implemented method demonstrated state-of-the-art results on KITTI dataset among unsupervised methods and showed robust performance on EuRoC. To the best of our knowledge, our visual odometry method is a pioneer work to train deep learning-base model on EuRoC in an unsupervised mode.

\newpage

{\small
\bibliographystyle{ieee}
\bibliography{references}

\begin{thebibliography}{10}\itemsep=-1pt

\bibitem{almalioglu2018ganvo}
Y.~Almalioglu, M.~R.~U. Saputra, P.~P. de~Gusmao, A.~Markham, and N.~Trigoni.
\newblock Ganvo: Unsupervised deep monocular visual odometry and depth
  estimation with generative adversarial networks.
\newblock {\em arXiv preprint arXiv:1809.05786}, 2018.

\bibitem{bian2019sc-sfmlearner}
J.-W. Bian, Z.~Li, N.~Wang, H.~Zhan, C.~Shen, M.-M. Cheng, and R.~I.
\newblock Unsupervised scale-consistent depth and ego-motion learning from
  monocular video.
\newblock {\em arXiv preprint arXiv:1908.10553}, 2019.

\bibitem{blanco2009cor}
J.-L. Blanco, F.-A. Moreno, and J.~Gonz{\'a}lez.
\newblock A collection of outdoor robotic datasets with centimeter-accuracy
  ground truth.
\newblock {\em Autonomous Robots}, 27(4):327--351, November 2009.

\bibitem{opencv_library}
G.~Bradski.
\newblock {The OpenCV Library}.
\newblock {\em Dr. Dobb's Journal of Software Tools}, 2000.

\bibitem{Butler:ECCV:2012}
D.~J. Butler, J.~Wulff, G.~B. Stanley, and M.~J. Black.
\newblock A naturalistic open source movie for optical flow evaluation.
\newblock In {A. Fitzgibbon et al. (Eds.)}, editor, {\em European Conf. on
  Computer Vision (ECCV)}, Part IV, LNCS 7577, pages 611--625. Springer-Verlag,
  Oct. 2012.

\bibitem{Claraco2010Tutorial}
J.~L.~B. Claraco.
\newblock A tutorial on se3 transformation parameterizations and on-manifold
  optimization.
\newblock Technical Report 012010, 2010.

\bibitem{costante2018ls}
G.~Costante and T.~A. Ciarfuglia.
\newblock Ls-vo: Learning dense optical subspace for robust visual odometry
  estimation.
\newblock {\em IEEE Robotics and Automation Letters}, 3(3):1735--1742, 2018.

\bibitem{dharmasiri2018eng}
T.~Dharmasiri, A.~Spek, and T.~Drummond.
\newblock Eng: End-to-end neural geometry for robust depth and pose estimation
  using cnns.
\newblock {\em arXiv preprint arXiv:1807.05705}, 2018.

\bibitem{dosovitskiy2015flownet}
A.~Dosovitskiy, P.~Fischer, E.~Ilg, P.~Hausser, C.~Hazirbas, V.~Golkov, P.~Van
  Der~Smagt, D.~Cremers, and T.~Brox.
\newblock Flownet: Learning optical flow with convolutional networks.
\newblock In {\em Proceedings of the IEEE International Conference on Computer
  Vision}, pages 2758--2766, 2015.

\bibitem{engel2018direct}
J.~Engel, V.~Koltun, and D.~Cremers.
\newblock Direct sparse odometry.
\newblock {\em IEEE transactions on pattern analysis and machine intelligence},
  40(3):611--625, 2018.

\bibitem{Feng2019sganvo}
T.~Feng and D.~Gu.
\newblock {SGANVO:} unsupervised deep visual odometry and depth estimation with
  stacked generative adversarial networks.
\newblock {\em CoRR}, abs/1906.08889, 2019.

\bibitem{Geiger2012CVPR}
A.~Geiger, P.~Lenz, and R.~Urtasun.
\newblock Are we ready for autonomous driving? the kitti vision benchmark
  suite.
\newblock In {\em Conference on Computer Vision and Pattern Recognition
  (CVPR)}, 2012.

\bibitem{DBLP:journals/corr/abs-1901-07288}
M.~Geng, S.~Shang, B.~Ding, H.~Wang, P.~Zhang, and L.~Zhang.
\newblock Unsupervised learning-based depth estimation aided visual {SLAM}
  approach.
\newblock {\em CoRR}, abs/1901.07288, 2019.

\bibitem{Henriques2018MapNet}
J.~F. Henriques and A.~Vedaldi.
\newblock Mapnet: An allocentric spatial memory for mapping environments.
\newblock In {\em The IEEE Conference on Computer Vision and Pattern
  Recognition (CVPR)}, June 2018.

\bibitem{kerl2013robust}
C.~Kerl, J.~Sturm, and D.~Cremers.
\newblock Robust odometry estimation for rgb-d cameras.
\newblock In {\em Robotics and Automation (ICRA), 2013 IEEE International
  Conference on}, pages 3748--3754. IEEE, 2013.

\bibitem{kuemmerle2011g2o}
R.~Kuemmerle, G.~Grisetti, H.~Strasdat, K.~Konolige, and W.~Burgard.
\newblock g2o: A general framework for graph optimization.
\newblock In {\em Proceedings of the IEEE International Conference on Robotics
  and Automation (ICRA)}, pages 3607--3613, Shanghai, China, May 2011.

\bibitem{Li2017UnDeepVO}
R.~Li, S.~Wang, Z.~Long, and D.~Gu.
\newblock Undeepvo: Monocular visual odometry through unsupervised deep
  learning.
\newblock {\em CoRR}, abs/1709.06841, 2017.

\bibitem{Lowe2004Test}
D.~G. Lowe.
\newblock Distinctive image features from scale-invariant keypoints.
\newblock {\em International Journal of Computer Vision}, 60(2):91--110, Nov
  2004.

\bibitem{Lowe1999Sift}
D.~G. Lowe, D.~G. Lowe, and D.~G. Lowe.
\newblock Object recognition from local scale-invariant features.
\newblock In {\em Proceedings of the International Conference on Computer
  Vision-Volume 2 - Volume 2}, ICCV '99, pages 1150--, Washington, DC, USA,
  1999. IEEE Computer Society.

\bibitem{Lv18eccv}
Z.~Lv, K.~Kim, A.~Troccoli, D.~Sun, J.~Rehg, and J.~Kautz.
\newblock Learning rigidity in dynamic scenes with a moving camera for 3d
  motion field estimation.
\newblock In {\em ECCV}, 2018.

\bibitem{mahjourian2018unsupervised}
R.~Mahjourian, M.~Wicke, and A.~Angelova.
\newblock Unsupervised learning of depth and ego-motion from monocular video
  using 3d geometric constraints.
\newblock In {\em Proceedings of the IEEE Conference on Computer Vision and
  Pattern Recognition}, pages 5667--5675, 2018.

\bibitem{MIFDB16}
N.~Mayer, E.~Ilg, P.~H{\"a}usser, P.~Fischer, D.~Cremers, A.~Dosovitskiy, and
  T.~Brox.
\newblock A large dataset to train convolutional networks for disparity,
  optical flow, and scene flow estimation.
\newblock In {\em IEEE International Conference on Computer Vision and Pattern
  Recognition (CVPR)}, 2016.
\newblock arXiv:1512.02134.

\bibitem{Menze2018JPRS}
M.~Menze, C.~Heipke, and A.~Geiger.
\newblock Object scene flow.
\newblock {\em ISPRS Journal of Photogrammetry and Remote Sensing (JPRS)},
  2018.

\bibitem{mur2017orb}
R.~Mur-Artal and J.~D. Tard{\'o}s.
\newblock Orb-slam2: An open-source slam system for monocular, stereo, and
  rgb-d cameras.
\newblock {\em IEEE Transactions on Robotics}, 33(5):1255--1262, 2017.

\bibitem{Schops2019badslam}
T.~Schops, T.~Sattler, and M.~Pollefeys.
\newblock Bad slam: Bundle adjusted direct rgb-d slam.
\newblock In {\em The IEEE Conference on Computer Vision and Pattern
  Recognition (CVPR)}, June 2019.

\bibitem{Slinko2019MotionMap}
I.~Slinko, A.~Vorontsova, F.~Konokhov, O.~Barinova, and A.~Konushin.
\newblock Scene motion decomposition for learnable visual odometry.
\newblock {\em CoRR}, abs/1907.07227, 2019.

\bibitem{steinbrucker2011real}
F.~Steinbr{\"u}cker, J.~Sturm, and D.~Cremers.
\newblock Real-time visual odometry from dense rgb-d images.
\newblock In {\em Computer Vision Workshops (ICCV Workshops), 2011 IEEE
  International Conference on}, pages 719--722. IEEE, 2011.

\bibitem{sun2018pwc}
D.~Sun, X.~Yang, M.-Y. Liu, and J.~Kautz.
\newblock Pwc-net: Cnns for optical flow using pyramid, warping, and cost
  volume.
\newblock In {\em Proceedings of the IEEE Conference on Computer Vision and
  Pattern Recognition}, pages 8934--8943, 2018.

\bibitem{Teed2019DeepV2D}
Z.~Teed and J.~Deng.
\newblock Deepv2d: Video to depth with differentiable structure from motion.
\newblock {\em CoRR}, abs/1812.04605, 2018.

\bibitem{ummenhofer2017demon}
B.~Ummenhofer, H.~Zhou, J.~Uhrig, N.~Mayer, E.~Ilg, A.~Dosovitskiy, and
  T.~Brox.
\newblock Demon: Depth and motion network for learning monocular stereo.
\newblock In {\em IEEE Conference on computer vision and pattern recognition
  (CVPR)}, volume~5, page~6, 2017.

\bibitem{DBLP:journals/corr/Vijayanarasimhan17}
S.~Vijayanarasimhan, S.~Ricco, C.~Schmid, R.~Sukthankar, and K.~Fragkiadaki.
\newblock Sfm-net: Learning of structure and motion from video.
\newblock {\em CoRR}, abs/1704.07804, 2017.

\bibitem{wang2017deepvo}
S.~Wang, R.~Clark, H.~Wen, and N.~Trigoni.
\newblock Deepvo: Towards end-to-end visual odometry with deep recurrent
  convolutional neural networks.
\newblock In {\em Robotics and Automation (ICRA), 2017 IEEE International
  Conference on}, pages 2043--2050. IEEE, 2017.

\bibitem{Wang2018Espvo}
S.~Wang, R.~Clark, H.~Wen, and N.~Trigoni.
\newblock End-to-end, sequence-to-sequence probabilistic visual odometry
  through deep neural networks.
\newblock {\em The International Journal of Robotics Research},
  37(4-5):513--542, 2018.

\bibitem{wang2018end}
S.~Wang, R.~Clark, H.~Wen, and N.~Trigoni.
\newblock End-to-end, sequence-to-sequence probabilistic visual odometry
  through deep neural networks.
\newblock {\em The International Journal of Robotics Research},
  37(4-5):513--542, 2018.

\bibitem{Xue019BeyondTracking}
F.~Xue, X.~Wang, S.~Li, Q.~Wang, J.~Wang, and H.~Zha.
\newblock Beyond tracking: Selecting memory and refining poses for deep visual
  odometry.
\newblock In {\em The IEEE Conference on Computer Vision and Pattern
  Recognition (CVPR)}, June 2019.

\bibitem{yin2018geonet}
Z.~Yin and J.~Shi.
\newblock Geonet: Unsupervised learning of dense depth, optical flow and camera
  pose.
\newblock In {\em Proceedings of the IEEE Conference on Computer Vision and
  Pattern Recognition (CVPR)}, volume~2, 2018.

\bibitem{ZbontarL2015disparity}
J.~Zbontar and Y.~LeCun.
\newblock Stereo matching by training a convolutional neural network to compare
  image patches.
\newblock {\em CoRR}, abs/1510.05970, 2015.

\bibitem{Zhan2018Depth-VO-Feat}
H.~Zhan, R.~Garg, C.~S. Weerasekera, K.~Li, H.~Agarwal, and I.~D. Reid.
\newblock Unsupervised learning of monocular depth estimation and visual
  odometry with deep feature reconstruction.
\newblock {\em CoRR}, abs/1803.03893, 2018.

\bibitem{Zhan2019learnt}
H.~Zhan, C.~S. Weerasekera, J.~Bian, and I.~Reid.
\newblock Visual odometry revisited: What should be learnt?
\newblock {\em arXiv preprint arXiv:1909.09803}, 2019.

\bibitem{zhao2018learning}
C.~Zhao, L.~Sun, P.~Purkait, T.~Duckett, and R.~Stolkin.
\newblock Learning monocular visual odometry with dense 3d mapping from dense
  3d flow.
\newblock {\em Intelligent Robots and Systems (IROS), 2018 International
  Conference on}, 2018.

\bibitem{zhou2018deeptam}
H.~Zhou, B.~Ummenhofer, and T.~Brox.
\newblock Deeptam: Deep tracking and mapping.
\newblock In {\em European Conference on Computer Vision (ECCV)}, 2018.

\bibitem{zhou2017unsupervised}
T.~Zhou, M.~Brown, N.~Snavely, and D.~G. Lowe.
\newblock Unsupervised learning of depth and ego-motion from video.
\newblock In {\em CVPR}, volume~2, page~7, 2017.

\end{thebibliography}
}
\clearpage

\includepdf[pages={1}]{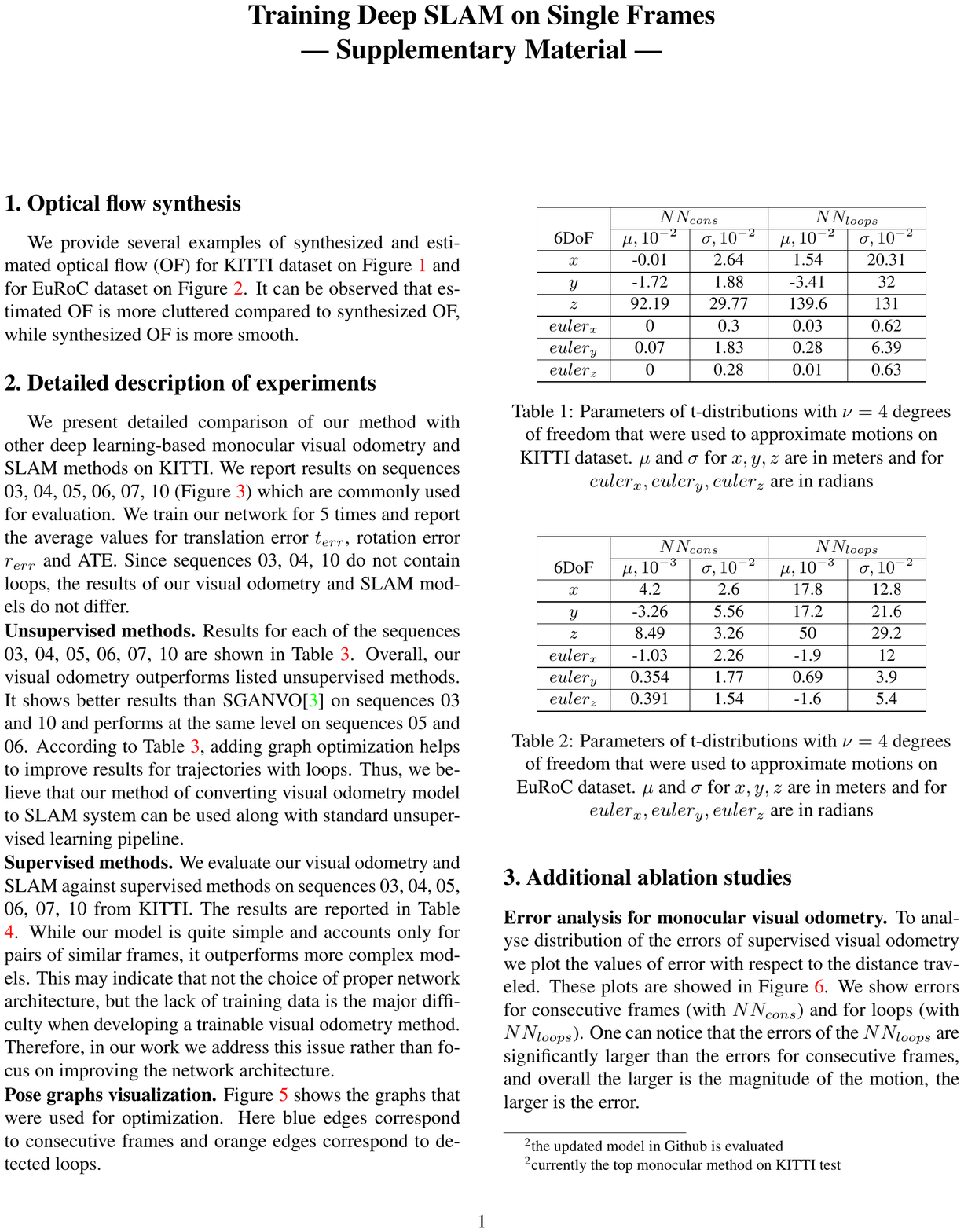}
\includepdf[pages={2}]{arXiv_DeepSLAM_supplementary.pdf}
\includepdf[pages={3}]{arXiv_DeepSLAM_supplementary.pdf}
\includepdf[pages={4}]{arXiv_DeepSLAM_supplementary.pdf}
\includepdf[pages={5}]{arXiv_DeepSLAM_supplementary.pdf}
\includepdf[pages={6}]{arXiv_DeepSLAM_supplementary.pdf}
\includepdf[pages={7}]{arXiv_DeepSLAM_supplementary.pdf}
\includepdf[pages={8}]{arXiv_DeepSLAM_supplementary.pdf}
\includepdf[pages={9}]{arXiv_DeepSLAM_supplementary.pdf}

\end{document}